# A New Framework for Machine Intelligence: Concepts and Prototype

**Abel Torres Montoya**
DataVeras
abel@dataveras.com

**Abstract**

Machine learning (ML) and artificial intelligence (AI) have become hot topics in many information processing areas, from chatbots to scientific data analysis. At the same time, there is uncertainty about the possibility of extending predominant ML technologies to become general solutions with continuous learning capabilities. Here, a simple, yet comprehensive, theoretical framework for intelligent systems is presented. A combination of Mirror Compositional Representations (MCR) and a Solution-Critic Loop (SCL) is proposed as a generic approach for different types of problems. A prototype implementation is presented for document comparison using English Wikipedia corpus.

## 1  Introduction

**The Key to the Future**  In current society, the amount of data and complexity of technology is continuously increasing; becoming too big to be processed directly by people. Equally, we are facing information processing challenges in many fields, e.g.: how to develop large, secure and stable software systems; how to extract the right information from the huge amount of cumulated cultural and technical knowledge; and how to productively analyze large scientific datasets. In this scenario, machine learning and artificial intelligence are more than good to have tools: they are required for building more advanced social, technological and scientific solutions. Future intelligent systems will be in charge of complex processes, performing continuous information management, research and anomaly detection.

**Path to General Solutions**  Currently, machine learning solutions based on neural networks are the reference tool in tasks like classification, object recognition and language translation. Deep learning (DL) is one of the most widely used implementations achieving consistent results based on the training data [1]. On the other hand, nowadays there is an increasing focus, from different perspectives, on the limitations of DL solutions in the path to general intelligent systems [2, 3, 4, 5]. Some of those limitations are: poor interpretability, need of a large amount of labeled data, poor controllability, maintainability, scalability, limited knowledge transfer and lack of abstract knowledge extraction, among others.

Several groups are working in modified versions of the deep neural networks to improve interpretability [6, 7, 8], visual reasoning [9], zero-shot learning [10, 11], intuitive physics [12, 13, 14, 15, 16], causality [17], catastrophic forgetting [18] and other issues.

Some of the multiple techniques being used include: attention [19], hierarchical models [20, 21], competing networks [22], autoencoders [23] and more. This variety of techniques raises the question on whether the path to overall performant solutions could be by putting all those techniques together on top of the deep network architecture.

In that sense, I share the conceptual approach taken in the capsules proposal [24] looking for a model that incorporates all main desired features in its design.

The framework proposed here is not based on DL architecture but takes into account the main features highlighted by improved DL models: compositionality, abstraction, hierarchical representations, collaborating networks, attention and continuous learning. The model uses explicit compositional representations to build complex abstractions from simple ones what seems to match neuroscience evidence [25, 26, 27]. The compositional structure facilitates interpretability, which is relevant beyond the common association with trustworthiness [28]. Interpretability is also important for maintainability and scalability of large complex systems.

## 2 Hierarchical Compositional Representations

**Decoupling Concepts** When can we say that a machine learning system is intelligent? This is a broad topic where many answers are available. Within the scope of intelligence as a problem solving capability, let's look for a practical definition that could be helpful for the implementation of such systems.

The following definition from Tom Mitchel is frequently cited [29]:

*"A computer program is said to learn from experience E with respect to some class of tasks T and performance measure P if its performance at tasks in T, as measured by P, improves with experience E"*

In this definition learning and intelligent processing appear together. This is also true in DL implementations: the system is designed for optimal task performance on the given training data. As result, *the part of the input not related to the task is discarded during the training*; causing a need to retrain with same inputs for another task, and eventually, leading to changes in the existing optimal representation for the first task.

To avoid this problem, in this proposal, learning (knowledge acquisition) and intelligent processing (goal satisfaction) are decoupled, creating a shared knowledge base which should contain all necessary information about the input to be used by different tasks.

> Note: In that sense, learning is assumed in the traditional educational sense of information storage and retrieval, where students are asked to retain information which is supposed to be useful later in life. Often, the learnt information is never used; which is part of the burden of a general intelligence.

Next, we need to decide which data representation to use for the knowledge base, and how to learn from an input in the absence of a task.

**Compositional World** People perceive, describe and explain the world as compositional with interacting abstracted elements; representing such process is an important topic in ML [30]. Explicit compositional representations would be natural, simple and sufficient way to store and retrieve information. Neuroscience evidence also supports the abstract representation approach with findings of activations of single neurons or small groups of neurons matching the presence of patterns, faces, places and other high level abstract concepts [31, 32]. Finally, aiming at high level cognitive functions: can a system reason by manipulating concepts without having an explicit representations of those concepts? Can a system execute a set of actions in arbitrary order without having and explicit representation of those actions?

Based on these arguments, *knowledge is chosen to be represented in a compositional hierarchical way* where every element is represented as a collection of lower level ones. In a simplified way, we could consider that there is a single top level node in the network for each concept (dog, bus, chair, etc.) defined by their subparts.

## 3 Learning with Mirror Compositional Representations

Keeping in line with the idea of direct correlation between perception and representation, we define the following learning criteria: *a good representation is one that can be used to replicate the input*. So, the goal of the training or learning in this model is matching the input using the existing set of learned patterns or adding new ones if needed. This is an unsupervised learning approach which offers better transparency during learning and data processing, as we can inspect what the system is 'thinking' by monitoring the activations of high level concepts. In this model there is no separate training phase and a continuous learning process takes place with each input.

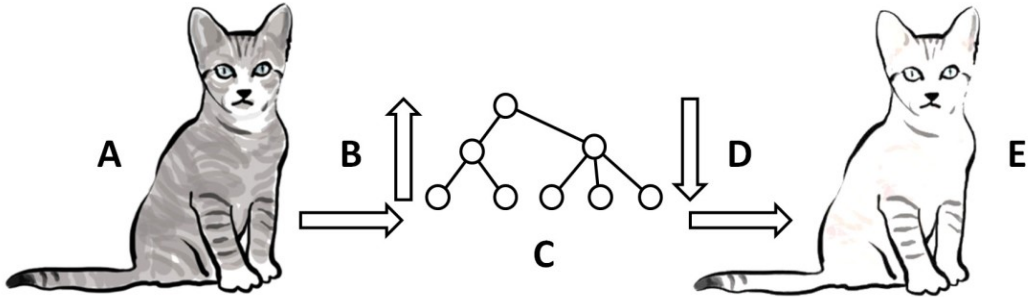

Figure 1: Replication as the goal of learning: A) Input, B) Bottom up unsupervised learning, C) Mirror compositional representation, D) Top down generative process, E) Output

The goal of learning is to detect all patterns of interest in the input and to incorporate them to the system. The internal representation C contains a sketched view E of the input A (Figure 1) depending on the amount of details needed by the system. MCR keeps the coherent structure of information at each abstraction level what should help reducing the impact of adversarial attacks based on using few lower level features to create non-plausible high level conclusions [33].

In the scope of a neural network solution, the information association is updated by creating new links and nodes or changing existing activation links between nodes of the network. In a simple activity propagation model, the set of nodes, links and activation parameters define the information processing capabilities of the system. Consequently, we can say that *learning changes the information processing to take into account a new input*. On the other hand, *intelligence is the process of changing the information processing of the system to meet a goal*.

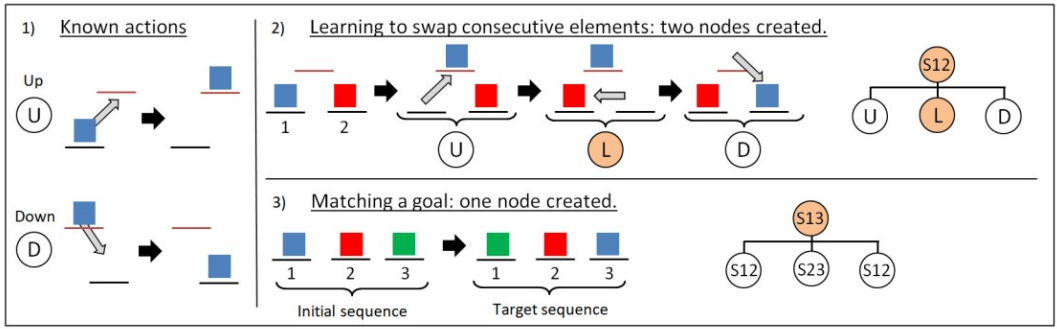

Figure 2: An object manipulation robot example. 1) Given a Knowledge Base containing two actions: U and D. 2) A new action sequence is shown to the system to be learnt: existing nodes are recognized and a new action L and top level node S12 are added. 3) During intelligent processing the input and target sequences are supplied and the system finds the combination of known actions to reach it, creating a new node S13.

## 4    Intelligence with Solution-Critic Loops

**Causal vs. Probabilistic**    To begin, we represent the goal internally in the system with a corresponding set of nodes. Using a means-ends approach, performing a task implies searching for the right high level nodes to be activated downstream matching goal's representation. This can be seen from the causal perspective as intelligence trying to explain top down the goal; while bottom up activation can be related to intuition where input is used to find a likely high level cause. In general, *intelligence uses a causal reasoning while intuition uses a probabilistic one*.

**Single Architecture**    In this framework, all information processing done by the system (goal satisfaction, causality search, understanding, imagination, etc.) is represented as a single iterative process of matching activation patterns of high level and low level nodes. Having a single architecture simplifies algorithm's implementation and corresponds to the expectation that biological intelligence uses a common rule. Also, such information processing loops are the norm in biological systems [34, 35].

**SCL** The general definition of intelligence for implementation purposes would be: *A system is intelligent if it contains an ad hoc process of generating solution candidates for a problem and criteria for evaluating them, favoring the better matches to the goal*.

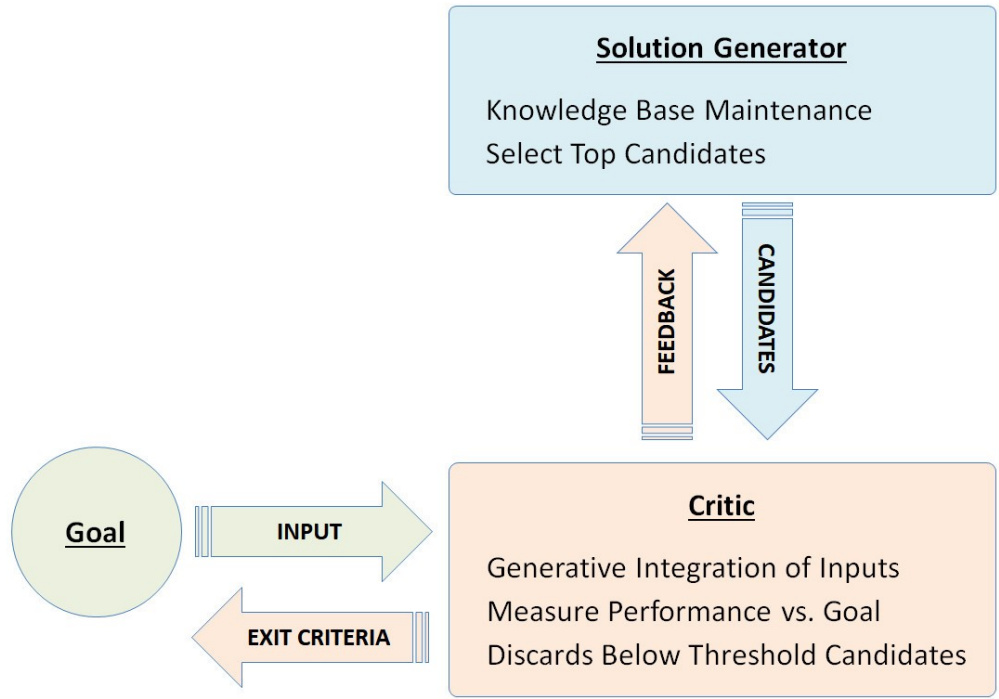

Figure 3: Intelligence as Solution-Critic Loop. The Solution Generator module creates proposals based on the feedback from the Critic module. The Critic computes the similarity between the consequences of the solution candidate and the goal. The loop continues until the given exit criteria are met (time-based, precision-based).

The SCL view of intelligence is applicable to biological systems where there is no abstract representation of the goal. This helps understanding how intelligence can solve problems just by using simple principles. In natural selection, the goal of organisms is to survive adverse conditions and the solution module consists of testing simultaneously many potential solutions by small changes in each member of the new generations. The verification and selection is done by natural selection and the DNA becomes the memory for next iterations. The approach of testing many solutions simultaneously is also taken by the slime mold to solve a maze. A more efficient approach should be used under time constraints, taking into account the feedback from the verification module to create new solution proposals.

Let's look at another popular definition, from S. Legg and M. Hutter [36]:

"*Intelligence measures an agent's ability to achieve goals in a wide range of environments*"

The contrast between this definition and the one based on dynamic loops can be illustrated in the following example. Let's assume that we know the optimal solution for each of multiple environments, and that we have a first, feedforward system, consisting of a detector, that identifies the right environment, and a set of links to the corresponding optimal solution. Let's consider also a second system which has no prior knowledge of the optimal solutions and which creates multiple candidates of solutions, comparing their performances. According to the proposed approach, the best performing system (first one) is not intelligent, while the second one, less performant, has an intelligent process for searching the solution. From a practical point of view, *intelligence is the process that helps in dealing with conditions that do not match any of our previous knowledge and can be measured by its capacity to offer alternative solutions to a problem*.

From this we can conclude that:

- Systems which have hardcoded robust and satisfactory solutions for the full range of environments do not need intelligence and will perform better than intelligent systems with partial knowledge of the scope
- Intelligence is a process which offers the advantage of general applicability but in order to be effective it needs:
    - A. An extensive compositional knowledge base containing the information learnt plus the cumulated solutions; and preferably, such that
        1. The solution to the problem can be found in the combinatorial space of the subparts
    - B. A strategy to increase his knowledge base until condition A.1 is met
    - C. An efficient approach to search the extensive space of potential solutions
    - D. A good mechanism for evaluating the performance of the solution candidates vs. the goal

How this approach relates to classic deep neural networks (DNN)?

Since, after training, the DNN behaves as an input-output mechanism, it is not intelligent at that point and is using only the hardcoded intelligent decisions adjusted for the training data. It is clear that only in very constrained conditions we can cover all configurations in a precomputed fashion; meanwhile, intelligent systems deal with the uncertainty attempting to dynamically populate the space around the specific task/environment configuration. This fact points to another of the limitations of DNN for general solutions.

> Note: The bottom-up feedforward activation of a DNN can be compared to pondering how much evidence we have for a given conclusion. This resembles human intuition and we could hypothesize that intuition is bottom up educated guess under uncertainty conditions in which intelligent loop cannot be engaged. According to this analogy, DNN algorithms create intuitive systems which perform non reasoned educated guesses.

## 5    Narrow and General Intelligence

As mentioned before, a very effective performance is not necessarily a sign of intelligence. *Intelligence is a measure of adaptability while performance is a measure of specialization*.

**Purpose of General Intelligence**   This leads us to the topic of relationship between narrow and general solutions. A data processing solution can be defined for a specific task or context (narrow) or to be adapted to many different tasks and contexts (general). The narrow solution can be more efficient than the general one as far as the conditions of the problem remain under certain scope. For a system that has to deal with a very large number of environments (e.g. unpredictable ones), it is more efficient to have a general algorithm. If we consider that narrow solutions do not appear suddenly, but evolve and improve from less specific ones, we can see general intelligence as the algorithm to discover and build narrow, efficient and specialized solutions to problems. In other words, the purpose of general intelligence is not, by default, to create the most efficient solution, *the purpose of general intelligence is just to find 'a solution', and if needed, to create an efficient narrow one*.

We can observe this while executing repetitive tasks; e.g. while learning a new motor skill, the intelligent process starts recruiting and tuning a group of nodes to execute the task; and after multiple iterations, the task is executed with precision without intelligence intervention. Intelligence acts as a critic for deciding on learning completion, after that, the resulting feedforward execution may be optimized discarding any non-essential structure (becoming black box like).

**Better than Human, Obviously**   As a side note: it should be expected that specialized machines perform better than humans at the optimized task, also in the scope of information processing. First, general intelligent systems learn a task under the 'good enough for the goal' criteria; second, the overall information processing loop (SCL) is always active becoming a distractor and consuming system's resources; and third, normalization of non-reasoning related properties (e.g. sensory and motor reaction times) would be necessary for some machine vs. human comparisons. There is no point in contrasting the performance of a narrow, specialized system, against another with general intelligence; the only thing that a general intelligent system can do to beat the narrow system is to create another, better, narrow solution. Taken outside their areas of specialization narrow systems fail in seemingly basic tasks.

## 6      Imagination and Subjective Experience

It has been said that prediction is the essence of intelligence and that our brains are prediction machines. Prediction is a relevant part of intelligent processing, as well as causal relationship and imagination. In the proposed hierarchical model, abstraction is crucial for hypothesis testing by combining different high level concepts into new solution candidates. Abstraction also allows moving in both directions in time, looking for causality or doing prediction. Then, in this model, *abstraction is the essence of general intelligence and brains are puzzle assembling machines* (at least, in what concerns rational behavior).

We can use a top down generative approach to forecast the action of the players in the given context. For example, while looking at an image of a lion and a gazelle in a wild life documentary, we can make a forecast of the next events based on the high level models created previously for that context. On the other hand, the full potential of the abstract representations is in integrating them in unseen scenarios (i.e. imagination). If now the gazelle and the lion are in a coffee shop they could take the same actions as in the savanna or just sit and order a tea.

Efficiently integrating multiple unrelated abstract elements is a challenging task. In biological intelligence such functionality seems to match the role of subjective experience: a mechanism to integrate all relevant information into one common scenario. Eventually (and maybe, not surprisingly), we may need subjective experience in systems with human level reasoning capabilities. Subjective experience may not be required if we have enough time and resources to explore the space of solutions using a good discrimination criteria before the scope of the problem expires. Intelligence does not require subjective experience, it needs, though, a good integrator and subjective experience is a great one.

In this model, the integration takes place at the Critic's module, separated from the solution generation. Interestingly, if we make a parallel between the Critic's module as the subjective experience and the Solution Generator as the subconscious, we can see some similitude with the existing evidence. We are exposed in our conscious perception to sudden, fully constructed ideas without being involved in their step by step elaboration; we can consciously influence the thinking process, extracting relations or using a clue to search for information which is, again, 'delivered' later. Finally, neuroscience experiments show a well measured delay between internal decision processes and conscious perception [37, 38].

## 7      Attention Mechanisms

As mentioned before, an intelligent system can be created with few basic components, but in order to achieve fast and accurate problem solving capabilities many optimizations may be required.

One of the key optimizations is the ad hoc context tuning, to constrain nodes activation to the current scenario. Context deals with the problem of many features in a general abstract representation by exposing only the relevant properties. By helping with focusing in particular parts of the knowledge base, context tuning plays the role of an attention mechanism, enabling task specific interpretation of the information. Focusing attention in a small area improves the computation speed at the cost of coverage.

In practice, all elements involved in the solution to the problem may be considered as part of the context and influence the parameters of the system. That includes the scenario, the task and even the system itself if we consider a shared knowledge base that can be used by many systems.

**Additional Mechanisms**      Other practical components of a functional data processing solution are: *monitoring* the activity of elements of interest and *rule enforcement* to change the behavior of the system when specific condition occurs (represented by a high level abstract node).

The representation of the SCL diagram on Figure 3 can be updated taking into account mentioned mechanisms (Figure 4).

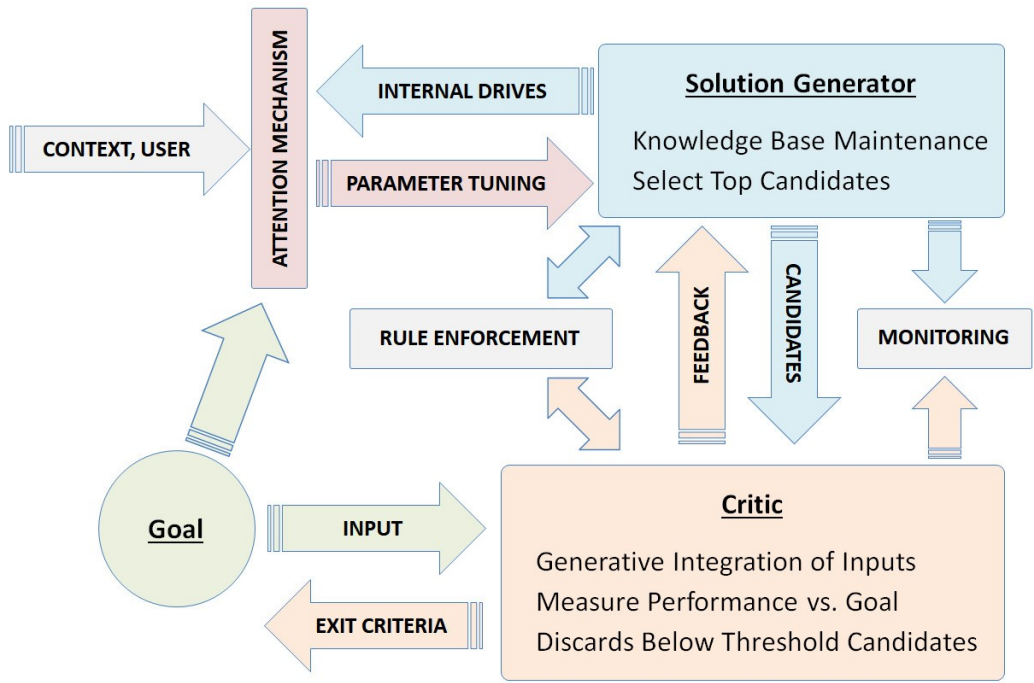

Figure 4: The intelligence framework including additional modules for attention, rule enforcement and monitoring. Many factors can influence attention: the goal, the context, the user input and the autoregulation of the system.

## 8  Document Comparison Prototype

Unstructured document comparison is at the heart of the intelligence algorithm. While in DNN we try to optimize a cost function to meet a goal, in this model, both the goal and the solution candidate are elements of the network. The possibility of specifying the goal in a 'natural' abstract representation seems as a big advantage, but it leads to the problem of measuring the similarity between two unstructured representations: goal and solution.

We compare unstructured data in a regular basis while recognizing a face or a place. It is used also to compare the similarity of financial trends or sequences of events. Imagine having two pictures where we see people, cars, trees and birds flying in the sky; all these elements in different amounts and perspectives. The question is: how can we measure the similarity between the two images?

**Bidirectional Activation**   To compute such metric we will calculate the 'influence' of each element on the other by propagating the activation top down from the first and bottom up to the second, and vice versa. Using this Bidirectional Activation (BA) approach we obtain two metrics, i.e., similarity is not a symmetrical property. For sake of simplicity we will combine the two values into a single metric.

Text document comparison is a particular application of the framework. In this case the knowledge base consists of a collection of documents to be searched for similarity to a given document.

**Content-Type Agnostic**   This prototype is using a collection of text documents but the algorithm is agnostic of the data type and can be applied to other hierarchical representations (objects, financial data, DNA sequences, etc.). No grammar or word meaning knowledge has been used in the implementation but such information could increase significantly the quality of the results and should be considered for a dedicated natural language processing task.  A hierarchical representation is created just based on words, phrases, sentences, paragraphs, chapters and articles.

Documents are a special case of unstructured data because language is already an abstraction of information. We cannot have a proper language understanding until the actual meaning of the words, their grammatical construction and, eventually, their context, are known. Nevertheless, in technical writing, articles tend to reflect the coherence of the elements they describe; this fact helps in the successful comparison using language agnostic mechanisms.

The Wikipedia corpus of 2015 (4.9 million articles) was selected as the knowledge base of the system. In a comparison task, top level articles are the potential solutions to approximate the given input document, which is playing the role of the goal. No combination of nodes is required for solution candidates in this task like in the general usage of the framework.

To propagate the activation in the system we need to assign some weights to the nodes. Those weights were calculated based on the overall frequency of the words in the knowledge base. The obtained values can be conditioned by the context. Two documents will compare differently depending on the active context (sports, food, culture, other). Here, no assumptions about the context are used to keep the comparison agnostic of the information type and to be aligned with the reference work by Dai, Olah and Le [39] used for the comparison. In their publication, the same Wikipedia version is used to find the closest articles to "Machine learning".

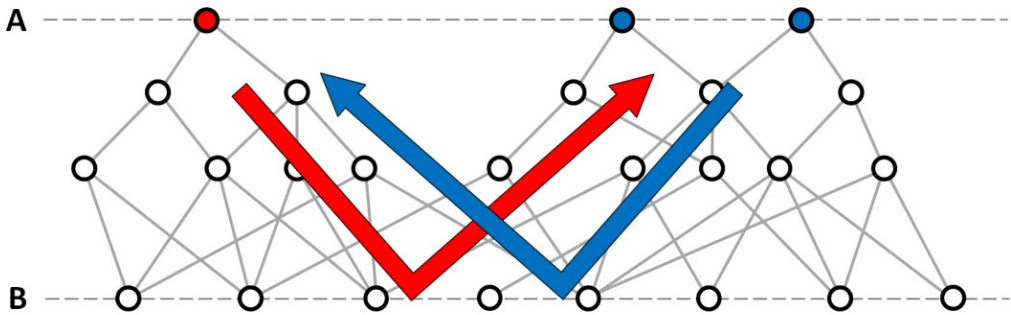

Figure 5. Bidirectional Activation between the goal (red) and the solutions (blue). The activation is propagated from the goal (red arrow) top down to the bottom layer B (words) and up to the topmost layer A (articles). Most active nodes at layer A are compared independently to the goal by reverse activation (blue arrow).

The activation is propagated from the target article to the database. The articles with highest activations are selected as 'solution candidates' and their influence on the target is computed. The two activations are combined with a simple formula: $S*ln(T)$, favoring the solution-to-target (S) activation value over the target-to-solution (T) one. To normalize the obtained values, we compute the final score as a percentage of the target-to-target activation.

| Paragraph Vector | Score(%) | Bidirectional Activation | Score(%) |
| --- | --- | --- | --- |
| **Artificial neural network** | | Semi-supervised learning | 28.6 |
| Types of artificial neural networks | 3.4 | Reinforcement learning | 22.7 |
| **Unsupervised learning** | | **Unsupervised learning** | **19.4** |
| Feature learning | 7.3 | Decision tree learning | 16.3 |
| Predictive analytics | 5.9 | Multi-task learning | 16.2 |
| Pattern recognition | 12.6 | **Artificial neural network** | **15.1** |
| Statistical classification | 6.0 | Computational learning theory | 14.7 |
| Structured prediction | 2.1 | K q-flats | 14.1 |
| Training set | 4.4 | Supervised learning | 13.9 |
| Meta learning (computer science) | 1.8 | Ensemble learning | 13.6 |

Table 1. Closest Wikipedia articles to "Machine learning". Comparison of the results using Paragraph Vectors and Bidirectional Activation; common articles are highlighted. All scores are from Bidirectional Activation (no comparison metric available in the reference article)

The inspection of the results for article comparison (Table 1) shows that the algorithm performs well selecting articles in the same topic as the target.

The goal of this prototype has been to create and test the basic mechanisms for unstructured data comparison. There are many possible improvements for the natural language content by incorporating context and grammar related information.

## 9 Discussion

A general problem solving framework, inspired by the compositional and creative properties associated to human learning and intelligence, is proposed based on two main elements:

- The Mirror Compositional Representation as a task-independent, hierarchical and unsupervised learning approach
- The Solution-Critic Loop as an ad hoc dynamic comparison of the high level representations of the goal and the solution

The practical application of the framework has been started with the implementation of the unstructured document comparison as a core feature of the SCL mechanism. The path forward includes performance improvement and extending the functionality to multiple data types. There is particular interest in testing the performance of the algorithm in combined data types and tasks beyond comparison and classification scenarios.

Although AGI is frequently associated with imitating human mind, here the focus is on developing intelligent systems with general capacity to solve data related problems (simulation, hypothesis discovery & verification, etc.) without aiming at imitating human behavior. Those intelligent systems will be capable of addressing the information processing challenges of large and complex processes.

This framework has multiple desired properties for large scale solutions:

- Transparency: the inspection of the activated abstract representations allows inspecting what the system is deciding at any time
- Interpretability: a trace can be created with all elements involved in a conclusion including the magnitude of their influence
- Controllability: the user can regulate the behavior of the system by modifying the activation patterns of the elements of interest (amplified, inhibited, etc.) as represented in the Rules Enforcement module (Figure 4)
- Maintainability/scalability: nodes can be added/replaced in a running system
- Knowledge extraction/transfer: new high level concepts are created by the system and knowledge trees can be transferred from one system to another
- Continuous learning: knowledge for different tasks can be isolated thanks to the attention mechanisms

Looking to the future:

- Single architecture solutions and rich goal representations will be critical for the generalization of ML solutions to many types of problems
- Knowledge extraction will be a bigger benefit from intelligent systems than task automation
- Abstract hierarchical models will be at the core of machine to human and machine to machine communication

## 10 Conclusions

Expectations are very high for machine learning systems following their recent progress. AI systems capable of dealing with bigger and more complex scenarios are needed for smart solutions in society, health care and science. Can we achieve such systems by increasing the size or partial modifications of current solutions or a paradigm change is required? This proposal wants to contribute to this discussion with a new promising approach.